%% file: main.tex
\documentclass{article}

     \PassOptionsToPackage{numbers, compress}{natbib}

\usepackage[preprint]{neurips_2021}

\usepackage[utf8]{inputenc} 
\usepackage[T1]{fontenc}    
\usepackage{hyperref}       
\usepackage{url}            
\usepackage{booktabs}       
\usepackage{amsfonts}       
\usepackage{nicefrac}       
\usepackage{microtype}      
\usepackage{xcolor}         
\usepackage{graphicx}
\usepackage{caption}
\usepackage{subcaption}
\usepackage{multirow}
\usepackage{algorithm}
\usepackage{algpseudocode}

\input{math_commands.tex}

\title{Sample-Efficient Generation of Novel Photo-acid Generator Molecules using a Deep Generative Model}

\author{%
  Samuel C.~Hoffman \\
  IBM Research\\
  Yorktown Heights, NY \\
  \texttt{shoffman@ibm.com} \\
   \And
   Vijil Chenthamarakshan \\
   IBM Research \\
   Yorktown Heights, NY \\
   \texttt{ecvijil@us.ibm.com} \\
   \AND
   Dmitry Yu.~Zubarev \\
   IBM Research \\
   Almaden, CA \\
   \texttt{dmitry.zubarev@ibm.com} \\
   \And
   Daniel P.~Sanders \\
   IBM Research \\
   Almaden, CA \\
   \texttt{dsand@us.ibm.com} \\
   \And
   Payel Das \\
   IBM Research \\
   Yorktown Heights, NY \\
   \texttt{daspa@us.ibm.com} \\
}

\begin{document}

\maketitle

\begin{abstract}
  Photo-acid generators (PAGs) are compounds that release acids ($H^+$ ions) when exposed to light. These compounds are critical components of the photolithography processes that are used in the manufacture of semiconductor logic and memory chips. The exponential increase in the demand for semiconductors has highlighted the need for discovering novel photo-acid generators. While de novo molecule design using deep generative models has been widely employed for drug discovery and material design, its application to the creation of novel photo-acid generators poses several unique challenges, such as lack of property labels. In this paper, we highlight these challenges and propose a generative modeling approach that utilizes conditional generation from a pre-trained deep autoencoder and expert-in-the-loop techniques. The validity of the proposed approach was evaluated with the help of subject matter experts, indicating the promise of such an approach for applications beyond the creation of novel photo-acid generators.
  
\end{abstract}

\section{Introduction}

De novo molecule design, the process of automatically generating molecules that satisfy multiple desirable characteristics, has been widely used in material design and drug discovery.
One area where new materials are urgently needed is semiconducting devices, which  are in ever increasing demand.  The recent shortage of microchips has demonstrated appreciable fragility of this critical technology under external perturbations.
For several decades, performance of semiconductor photolithography has been enabled by technology of chemical amplification \cite{PAG_History_1999, PAG_Design_2019}. The level of regulatory scrutiny of key components of chemical amplification, photo-acid generators, has been steadily increasing \cite{PAG_Env_2021, PAG_Env_2021A, PAG_Design_2011}. The consensus across the industry is that accelerated discovery of novel PAGs is the core strategy in meeting growing market demands while further advancing photolithography performance and achieving a high level of sustainability of semiconductor manufacturing.

Discovery of PAGs should take advantage of inexpensive yet informative metrics of photochemical performance along with established metrics of environmental impact and toxicity (ENV/TOX). Chemistry of photo-acid generation under photolithography conditions is rich and complex \cite{PAG_General_2020}. 
Some of the most important PAGs for industrial use are salts and comprise a photo-reactive cation from the family of sulfoniums, which is paired up with an anion to form a superacid.
Here, we focus on designing  novel sulfoniums that can exhibit favorable  photo-reactive properties, such as a low energy of the lowest occupied molecular orbital (LUMO) of the molecule, which can be estimated by quantum chemical calculations. LUMO energy provides a good first approximate of the  electron attachment (EA) energy of the photoactive component of a PAG.
Other desired properties of interest are high environmental sustainability and low toxicity.

The chemical space of  molecules is large ($10^{60}$) \cite{polishchuk2013estimation}. De novo design of molecules requires  traversing the  space of molecules  to efficiently search for the small number of molecules that might satisfy the desired criteria (in this case PAG cations  with low LUMO energy).
For many years, metaheuristic or gradient free methods like evolutionary algorithms and particle swarm optimization have been widely used for for this purpose \cite{reutlinger2014multi,yuan2011ligbuilder,nigam2019augmenting,winter2019efficient}. However, recently, deep learning (gradient based) approaches like Generative Adversarial Networks (GAN), Variational Autoencoders (VAE) have gained prominence. 

In this paper, we propose an attribute-controlled generation   approach that leverages a deep generative model to tackle the design of novel PAG cations for semiconductor industry application. Deep learning based approaches typically require a large amount of training data to learn a good representation of the underlying domain. However, we only have a few thousand known photoactive components of PAGs that are publicly available. To tackle this problem, we use a pre-training strategy that trains a deep generative autoencoder model on the large number of cationic molecules that are reported in chemical databases. Next, we employ conditional latent sampling to generate novel PAG cations, which leverages  a  latent classifier trained on a  small number of photo-acid generators. Generated molecules are evaluated and compared with the training data by using a number of established metrics.  We further evaluate the generated molecules with the help of subject matter experts and report the successes and avenues for further improvement.

\section{Related Work}
Reported studies dedicated to the design of PAGs for photochemical applications primarily followed \textit{ad hoc} approaches based on chemical intuition and design rules (cf. \cite{PAG_Design_2019, PAG_Design_2011}) coupled with pre-existing empirical knowledge of environmentally friendly molecular building blocks. Generative modeling of sulfoniums in the sense of contemporary deep learning approaches to small molecule discovery in drug design is yet to become a standard technique among practitioners of the art. Sulfonium cations comprise a relatively small set (approximately 1200 compounds in PubChem database if searched by "sulfonium" keyword). This suggests an enormous potential for their expansion and also hints at the potential challenges to the generative modeling. One challenge is relative scarcity of the constraints on sulfonium generation compared to the available constraints on the small molecule drug design. Access to the performance characteristics and ENV/TOX metrics is limited, as the application domain is siloed due to its niche nature and high requirements for intellectual property protection. Finally, the economy of this field of chemical research is not conducive to large-scale expensive experimentation and characterization, further limiting volumes and quality of the knowledge driving generative modeling. One practical strategy to resolve these challenges is to pair up generative modeling with expert-in-the-loop approaches \cite{EITL_2020}. Along that line, we introduced adjudication stage where a subject matter expert (SME) experienced in photolithograpy reviewed and labeled generated candidates.

Several generative modeling approaches that utilize deep learning models have been proposed for de novo generation of molecules. These include, but not limited to, Recurrent Neural Networks \cite{segler2017generating, gupta2018generative}, Generative Adversarial Networks (GAN) \cite{guimaraes2017objective, de2018molgan} and Variational Autoencoders (VAE) \cite{gomez2018automatic, blaschke2018application, kang2018conditional, lim2018molecular}.
Reinforcement Learning, Bayesian Optimization and semi-supervised learning based methods have been used for goal-directed  generation of molecules~\cite{  kang2018conditional, lim2018molecular, popova2018deep, olivecrona2017molecular,jaques2017sequence, putin2018reinforced, Zhavoronkov2019natbio, gomez2018automatic, born2020paccmann, zhou2019optimization,  li2018multi}.

\section{Model and Methods}
Our approach builds on prior work in conditional generation using attribute-guided  rejection sampling from a continuous latent space, namely Conditional Latent (attribute) Space Sampling (CLaSS) \cite{das2020science,chenthamarakshan2020cogmol}. CLaSS works by first training a generative  framework based on a deep latent variable model, such as a VAE, on a large set of unlabeled data.  Then, a density model of the learned latent variables is  constructed.
Attribute-conditioned generation is then performed by sampling from a desired conditional distribution, resulting in molecules with desired properties (in this case, cations with low LUMO energies). The conditional sampling  leverages a rejection sampling scheme from the models in the latent z-space appealing to Bayes rule, and attribute classifiers trained on the latent variables \cite{das2020science}. See Appendix \ref{sec:cond_sampling} for further details on conditional sampling.
We also draw inspiration from CogMol that utilizes CLaSS in conjunction with additional \textit{in silico} screening procedures \cite{chenthamarakshan2020cogmol} to reduce the pool of designs for further evaluation.

\subsection{Data}
\label{data}
We start with a set of onium salt PAG compounds compiled from patent data and split the compounds into the respective cations and anions. In this work, we focus on the cations only. We then compute the molecular orbitals with B3LYPV3/3-21G density functional theory (DFT) using GAMESS \cite{GAMESS} and save the highest occupied and lowest unoccupied molecular orbital (HOMO and LUMO) energies for each molecule. We end up with 1353 cations: this forms the basis of our PAG training data.

We also needed data on which to train the VAE model. For this, we used the ZINC15 dataset \cite{zinc} as it is large (over 700\,000 molecules) and similar in length and element type to the PAG training set. We further restrict to the samples from ZINC to those that are  positive ions and are within the minimum and maximum of the observed PAG samples in the following attribute values: number of atoms, hydrophobicity (logP), synthetic accessibility (SA), molecular weight (MW), number of rings, and maximum ring size. We also limit the ZINC molecules to only contain atoms seen in the PAG set. A summary of this process can be seen in Table \ref{tab:zinc_cations}. Roughly 163\,000 cations from ZINC remain after the filtering process, refered as the ZINC-filtered dataset.

\begin{table}[tb]
\centering
\caption{Description of ZINC dataset after filtering to match the minimum and maximum property values from PAG dataset. The final size of this ZINC-filtered dataset is around 163\,000 cations.}
\label{tab:zinc_cations}
\begin{tabular}{@{}l|rr@{}}
\toprule
               & minimum                  & maximum                   \\ \midrule
Vocab.         & \multicolumn{2}{c}{\{Br, C, Cl, F, I, N, O, S, Si\}} \\
Num. atoms     & 4                        & 79                        \\
logP           & $-$5.68                    & 23.73                     \\
SA             & 1.82                     & 7.91                      \\
Mol. weight    & 58.10                    & 984.18                    \\
Num. rings     & 0                        & 12                        \\
Max. ring size & 0                        & 6                         \\ \bottomrule
\end{tabular}
\end{table}

\subsection{Variational Autoencoder Training}

We use a bidirectional Gated Recurrent Unit (GRU) architecture for the encoder and decoder of the VAE with a latent dimension of 128. This is the same model architecture used in CogMol \cite{chenthamarakshan2020cogmol}. We add additional loss terms for predicting logP, SA, and Morgan fingerprints (FP) from the latent vectors in order to better regularize the latent space. These prediction models are simple 4-layer MLPs with hidden dimensions 50 and dropout probability 0.2 and are trained end-to-end with the VAE. SA and logP use L1 loss while FP uses cross entropy and logP is weighted at 0.1 times compared to SA and FP.

Following the approach of CLaSS, we fit a Gaussian mixture model to approximate the density of the latent features of all the PAG cations. We use 100 components and diagonal covariance matrices as well.

\begin{table}[tb]
\centering
\caption{Distance/similarity metrics with respect to known  sulfonium PAG reference set with low LUMO energies, as well as generation quality metrics. GEN is a set of 10\,000 molecules generated by our model and ZINC is the ZINC-filtered data described in Table \ref{tab:zinc_cations}. Refer to MOSES for description of metrics \cite{moses}. Better values are \textbf{bolded}.}
\label{tab:metrics}
\begin{tabular}{@{}l|rrrrrrr|rrr@{}}
\toprule
          & \multicolumn{7}{c|}{Distance/similarity}              & \multicolumn{3}{c}{Quality} \\ \midrule
Comp. set & FCD  & SNN   & Frag. & Scaf. & MW    & logP  & SA    & IntDiv & Uniq. & Nov. \\ \midrule
GEN       & \textbf{10.57} & \textbf{0.336} & \textbf{0.693} & \textbf{0.432} & \textbf{7102} & 3.04  & 0.122 & 0.843 & 0.969 & 0.993 \\
ZINC      & 21.53 & 0.225 & 0.538 & 0.035 & 13980 & \textbf{0.475} & \textbf{0.088} & 0.844 & -- & -- \\
\bottomrule
\end{tabular}
\end{table}

\subsection{Attribute Predictors}
\label{sec:attr}

As mentioned previously, we are interested in molecules with low LUMO energy. Specifically, we chose a threshold of $-5$ eV and lower. In order to perform rejection sampling on the VAE samples, we trained a binary classifier to predict high/low LUMO energy according to this threshold using the PAG data from Section \ref{data} using a simple 1 layer MLP with hidden dimension 100. This classifier achieves 79.4\% balanced accuracy using 5-fold cross-validation. Figure \ref{fig:lumo_conf_mat} shows the confusion matrix for this classifier.

We also incorporated additional filtering criteria based on feedback from expert chemists. This allows the outputs to be tuned for explicit, directly-computable properties that augment our definition of ``good'' PAGs. These included limiting generated molecules to only sulfonium ions and discarding amines and fluorine-rich molecules. The acidic properties of amines can interfere with photo-acid generation while fluorines have negative environmental impacts that we wish to avoid in de novo compounds.

\section{Results}

\begin{figure}[t]
    \centering
    \includegraphics[height=0.3\textwidth]{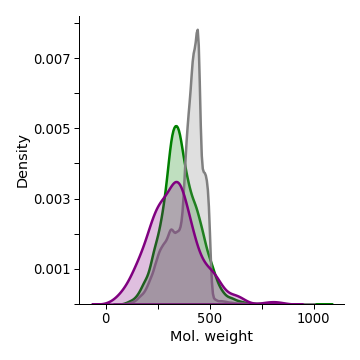}
    \includegraphics[height=0.3\textwidth]{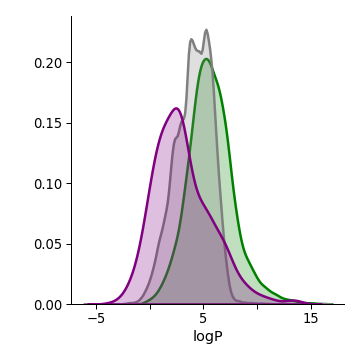}
    \includegraphics[height=0.3\textwidth]{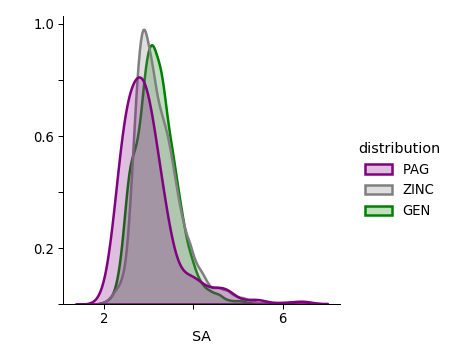}
    \includegraphics[height=0.3\textwidth]{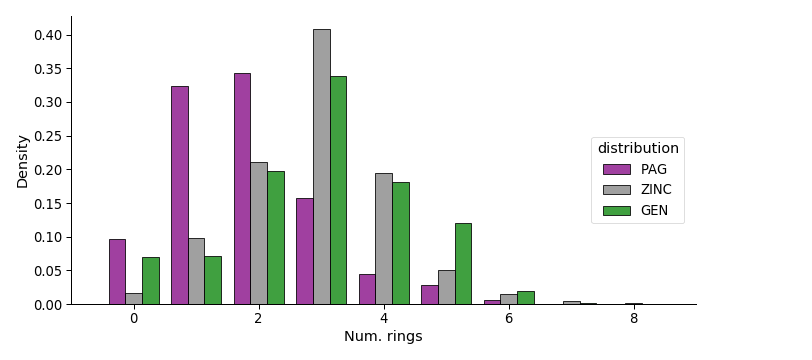}
    \hspace{-3em}
    \includegraphics[height=0.3\textwidth]{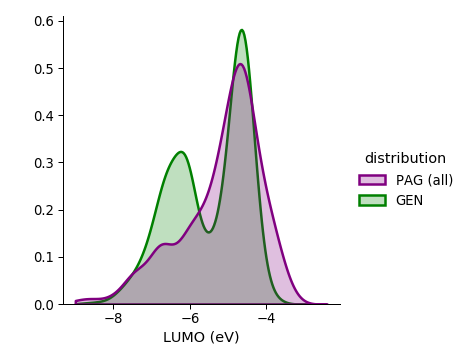}
    \caption{Distributions of various metrics for generated molecules (GEN) compared to low-LUMO, $<-5$ eV, known sulfonium PAGs and the filtered ZINC cations described in Table \ref{tab:zinc_cations}. The LUMO plot compares \textbf{all} PAGs to the generated molecules.}
    \label{fig:metric_dists}
\end{figure}

We evaluated the performance of our generation pipeline by comparing the distribution of generated molecules (after the additional filtering steps described in Section \ref{sec:attr}) to various reference sets. In Table \ref{tab:metrics}, we compute metrics from the MOSES benchmark \cite{moses}. For the distance/similarity metrics, generated molecules and ZINC-filtered molecules from the filtered ZINC training set are compared to the reference set which is the same as the PAG training data but limited to LUMO < $-5$ eV sulfonium cations. Internal diversity is also calculated for both comparison sets and uniqueness and novelty are calculated for the generated molecules to assess the quality of the generator. Overall, our generation pipeline succeeds in shifting the generated distribution of molecules from the pretrained distribution, closer to the desired PAG-like distribution while producing novel, unique, and diverse molecules, as indicated by lower  Fréchet ChemNet Distance (FCD) and higher Nearest neighbor similarity (SNN), fragment similarity (Frag.) and scaffold similarity (Scaf.). In contrast, the generated PAGs are subtantially different in terms of physical properties such as lipophilicity (logP) and synthetic accessibility (SA). 

The kernel density estimation plots in Figure \ref{fig:metric_dists} help further illustrating this point.
Interestingly, while  the generated PAGs do show a significant overlap with the logP, SA, and molecular weight (MW)  distributions of the training PAG molecules , there appears PAG generations that are physico-chemically distinct from the known PAGs. For example, generated PAGs do show a characteristic ring count distribution that is closer to the ZINC-filtered cations, when compared to known PAGs.
Furthermore, the LUMO distribution (as validated by DFT computation) shows that while many generated molecules remain near the threshold of $-5$ eV, there is a significant increase in low-LUMO molecules compared to the full PAG training data.

\begin{table}[t]
\centering
\caption{Summary of generated data evaluated by expert chemist. Reference is the data collected from patents while Generated is the samples from our model.}
\label{tab:scaf}
\begin{tabular}{@{}l|rr@{}}
\toprule
                          & Reference & Generated \\ \midrule
Sulfonium cations         & 4619      & 812       \\
All scaffolds             & 471       & 134       \\
Sulfonium scaffolds       & 255       & 69        \\
Novel sulfonium scaffolds &    --       & 61        \\ \bottomrule
\end{tabular}
\end{table}

Further evaluation of the ``goodness'' of generated PAGs poses a problem since this is a complex topic which cannot be distilled into a single, or even multiple, numerical property. In this work, we ask a subject matter expert (SME) chemist to evaluate the potential of generated scaffolds for photo-acid generation. Scaffolds were chosen for evaluation because they distill key traits of the molecule but remain easy to modify by chemists.

\begin{figure}[b]
    \centering
    \includegraphics[width=\textwidth]{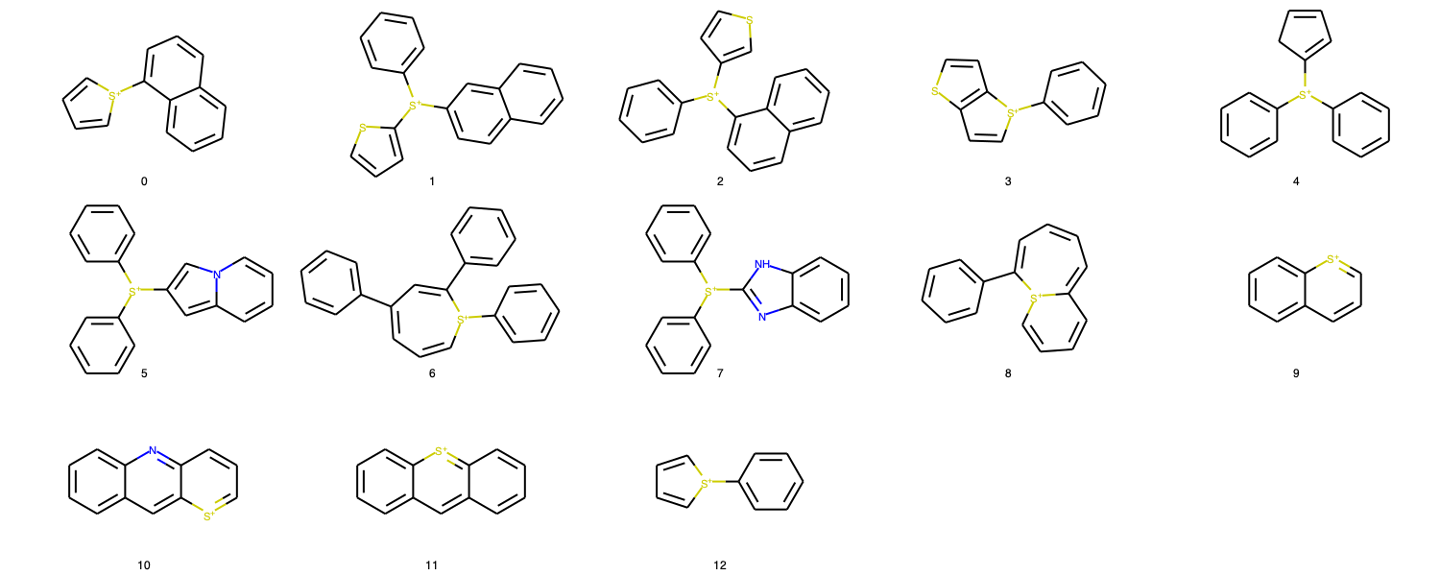}
    \caption{Generated scaffolds evaluated as promising by SME.}
    \label{fig:promising_scaffolds}
\end{figure}

We first sampled a large number of sulfonium cations using the pipeline described above. Then, we calculated the maximum fingerprint similarity with any of the samples from the known PAG cations set for each of the samples and selected at most 100 at random from each bin of 0.1 similarity increments. Generally patent-worthy sulfonium cations are dissimilar from each other (see Appendix \ref{sec:patent_dice}) so it makes sense to choose such molecules for analysis. On the other hand, more familiar molecules are easier to analyze and build confidence in the model. This left us with 812 molecules (none were found to be between 0 and 0.1 similarity and few were between 0.9 and 1; exact matches were discarded). From there, the generated cations were split into BRICS decomposition fragments \cite{degen2008} and Bemis-Murcko scaffolds were extracted from the fragments \cite{murckoscaf}. Finally, non-sulfonium scaffolds were removed from consideration. The results of this processing is summarized in Table \ref{tab:scaf}. The 61 resulting novel sulfonium scaffolds were evaluated by the SME and 13 promising candidates were identified. These can be seen in Figure \ref{fig:promising_scaffolds}. Notably, we observe approximately 2 orders of magnitude improvement in terms of viable candidates compared to a meta-heuristic based algorithm (unpublished results).

\begin{figure}[t]
    \centering
    \includegraphics[width=\textwidth]{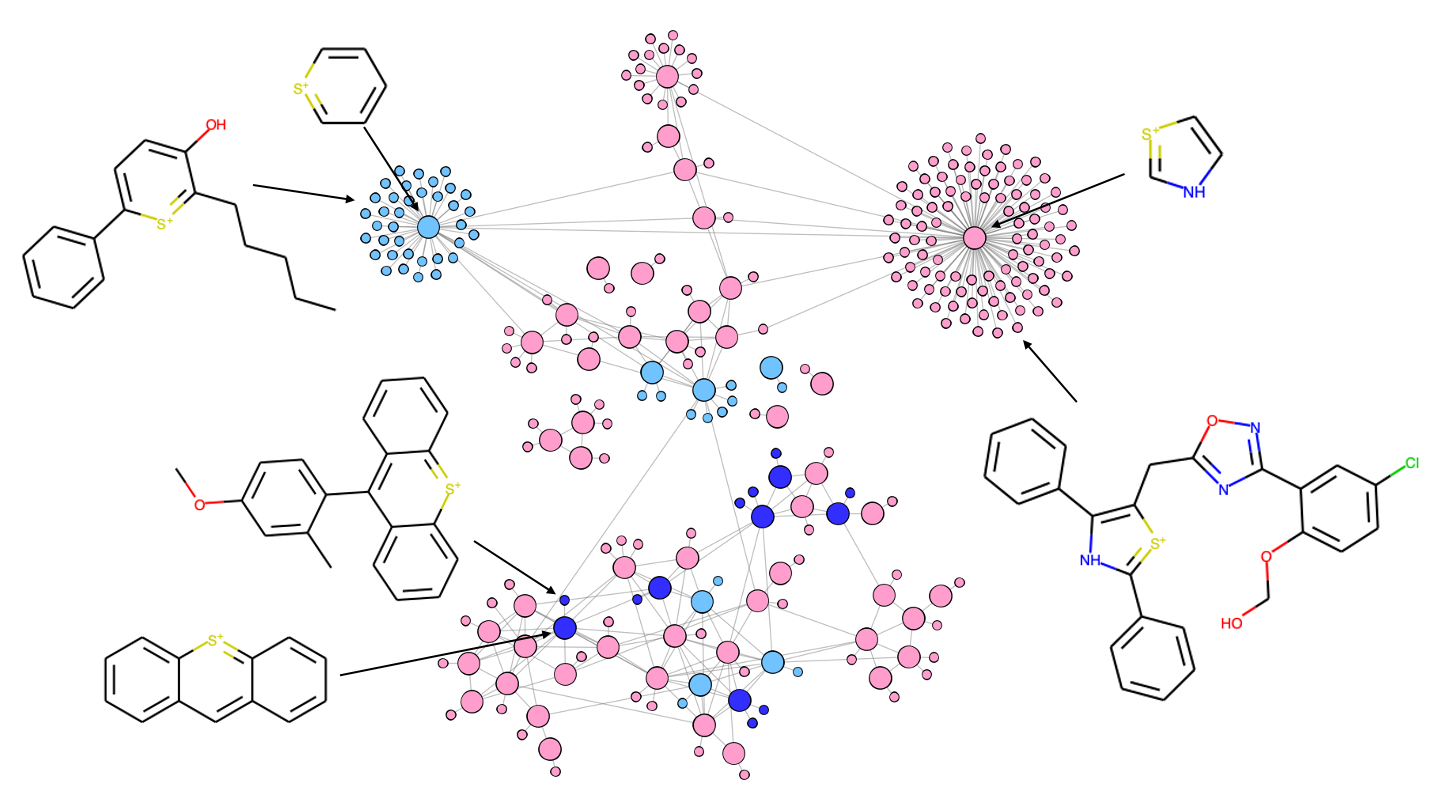}
    \caption{Network representation of the set of the adjudicated scaffolds (large nodes) and their parent molecules (small nodes). Color encodes SME decision about the scaffolds and is propagated to the parent molecules for visualization purposes: pink - rejection, light blue - uncertain, and deep blue - acceptance. Two scaffolds are linked if Dice distance between their Morgan fingerprints is less than 0.65. A molecule and a scaffold are linked if the scaffold is derived from the molecule.}
    \label{fig:scaff2mol_vis}
\end{figure}

\subsection{Characterization of Designed PAGs based on SME feedback}

The primary adjudication task of the SME is to reject non-viable candidates. The adjudication was carried out over scaffolds derived from the output of the generative model. The candidates were rejected if some of the aspects of their structure raised strong concerns about general stability under lithographic conditions, ability to release a proton as a result of photo-rearrangement, risk of quenching released protons due to the presence of basic functional groups, risk of side reactions due to the presence of reactive groups, presence of moieties that are highly toxic, environmentally harmful, or avoided by industrial vendors. Systematic enumeration of specific criteria that contributed to the SME's decisions is an ongoing effort that involves construction of a specialized ontology and remains outside the scope of the presented study. The scope and results of the adjudication are expressed as a network in Figure \ref{fig:scaff2mol_vis}.

The set of candidates that were not rejected was refined by asking the SME about the level of their confidence about viability of the candidates expressed as "accept" and "uncertain" labels. Further validation can then be performed on the adjudicated candidates. Of the candidates identified by our model, 8 were assessed to have meaningful synthetic routes and at least 1 satisfied stricter criteria of viability, function, ENV/TOX properties, and IP potential.

\section{Conclusions}
The presented approach successfully addresses the issue of a fundamentally small volume of preexisting structural data accessible to a generative model. The overall strategy pairs up the attribute-conditional sampling from a pre-trained deep generative model trained on a large amount of unlabeled data with SME adjudication. This is particularly attractive under the constraints of the discovery tasks, where appreciable design constraints, i.e. the definition of ``goodness'', is not available in explicit form, due to the combination of the siloed nature of the domain, lack of practices codifying the constraints, and varying priorities and preferences of different experts. In summary, we show that a deep generative modeling approach to de novo molecule design can be applied to great effect for generating novel photo-acid generator cations that are safe, effective, and sustainable and we hope that this work can serve as a blueprint for future works with similar challenges.

\bibliographystyle{IEEEtran}
\bibliography{confs}

\newpage
\appendix
\section{Appendix}

\subsection{Additional background on conditional sampling}
\label{sec:cond_sampling}

\begin{algorithm}[ht]
\caption{Conditional Latent (attribute) Space Sampling (CLaSS)}
\label{alg:class}
\begin{algorithmic}[1]
\Require Trained latent variable model (e.g.~VAE), samples $\rvz_j$ drawn from domain of interest, labeled samples for each attribute $a_i$.
\State Encode training data $\rvx_j$ in latent space: $\rvz_j \sim q_{\phi}(\rvz|\rvx_j)$
\State Use $\rvz_j$ to fit explicit density model $Q_{\varepsilon}(\rvz)$ to approximate marginal posterior $q_{\phi}(\rvz)$
\State Train classifier models $q_{\varepsilon}(a_i|\rvz)$ using labeled samples for each attribute $a_i$ to approximate probability $p(a_i|\rvx)$
\State Assuming attributes $a_i$ are conditionally independent given $\rvz$, then
    $$\hat{p_{\varepsilon}}(\rvz|\rva) = \frac{Q_{\varepsilon}(\rvz)\prod_i{q_{\varepsilon}(a_i|\rvz)}}{q_{\varepsilon}(\rvz)}$$
via Bayes' rule.
\State Let $g(\rvz) = Q_{\varepsilon}(\rvz)$ and $M = \frac{1}{q_{\varepsilon}(\rva)}$
\Repeat
    \State Sample from $Q_{\varepsilon}(\rvz)$
    \State Accept with probability $\frac{f(\rvz)}{Mg(\rvz)} = \prod_i{q_{\varepsilon}(a_i|\rvz)} \leq 1$
    \If{Accepted}
        \State Decode sample from latent and save: $\rvx \sim p_\theta(\rvx|\rvz)$
    \EndIf
\Until{Desired number of samples attained}
\State \Return Accepted samples
\end{algorithmic}
\end{algorithm}

We briefly describe Conditional Latent (attribute) Space Sampling (CLaSS) as proposed in \cite{das2020science}.
CLaSS uses attribute predictors trained on the latent embeddings of a Variational Autoencoder along with a density model to generate molecules through rejection sampling.
Let $\rva \in \R^n = [a_1, a_2, \dots, a_n]$, be a set of attributes of interest, which are assumed to be independent. CLaSS performs conditional sampling as $p(\rvx|\rva) = \E_\rvz[p(\rvz|\rva) p(\rvx|\rvz)]  \approx \E_\rvz[\hat{p}_\xi(\rvz|\rva) p_\theta(\rvx|\rvz)]$. The term $\hat{p}_\xi(\rvz|\rva)$ is approximated using a Gaussian Mixture Model $Q_\xi(\rvz)$  and per-attribute classifier model $q_\xi(a_i|\rvz)$ through Bayes rule. Rejection sampling is performed through the proposal distribution: $g(\rvz) = Q_\xi(\rvz)$.

\subsection{Attribute predictor details}
The confusion matrix for our attribute (LUMO) classifier is given below in Figure \ref{fig:lumo_conf_mat}. The attribute classifier is trained on the latent embeddings of the Variational Autoencoder (VAE).
\label{sec:conf_matrix}

\begin{figure}[b]
    \centering
    \includegraphics[width=0.5\textwidth]{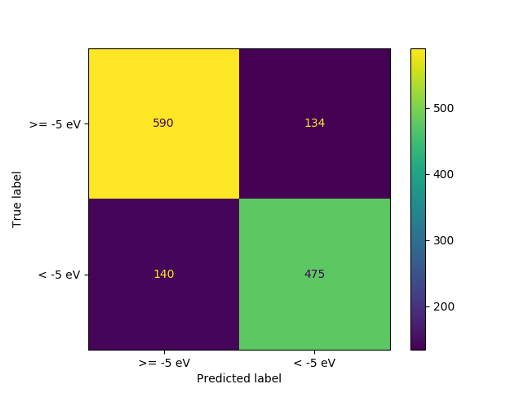}
    \caption{Confusion matrix for LUMO classifier.}
    \label{fig:lumo_conf_mat}
\end{figure}

\subsection{Analysis of patent-worthiness of sulfonium cations}

\label{sec:patent_dice}

\begin{figure}[t]
    \centering
    \includegraphics[width=0.6\textwidth]{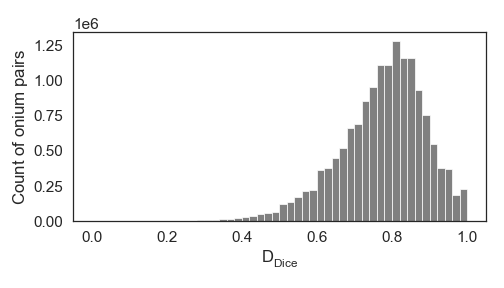}
    \caption{Histogram of pair-wise Dice distances over the set of sulfonium cations from patents. Mode at 0.8 suggests an empirical bound on how similar patent-worthy sulfoniums can be.}
    \label{fig:dice_dist}
\end{figure}

For a sulfonium cation to be a viable candidate for the development of intellectual property and have high patent-worthiness, it has to be both novel and structurally non-obvious at the time of the patent filing. Generation of sulfonium candidates with significantly different structures should be prioritized over incrementally changing structures. This strategy is substantiated by the analysis of the histogram of pair-wise Dice distances on the Morgan fingerprints of the sulfonium cations present in the patents shown in Figure \ref{fig:dice_dist}. The mode of the histogram corresponds to the Dice distance 0.8, indicating that the data set of patent-worthy sulfoniums comprises predominantly dissimilar structures and offering a simple operational measure of patent-worthiness that can be used to inform molecular generation process. 

\subsection{LUMO energy of SME-evaluated generated sulfonium cations}
\label{sec:accepted_lumo}

Generated sulfonium cations corresponding to the scaffolds evaluated by the SME as promising and ``accepted'' are shown below in Figure \ref{fig:accepted_lumo} along with the LUMO energy computed with DFT. The average LUMO energy for this set was $-5.45$ eV.

Another set of scaffolds derived from the generated molecules were evaluated as ``uncertain'' but still possibly promising. The parent molecules corresponding to these scaffolds are also shown below in Figure \ref{fig:uncertain_lumo}. The average LUMO energy computed for this set was $-6.57$ eV.

\begin{figure}[b]
    \centering
    \includegraphics[width=\textwidth]{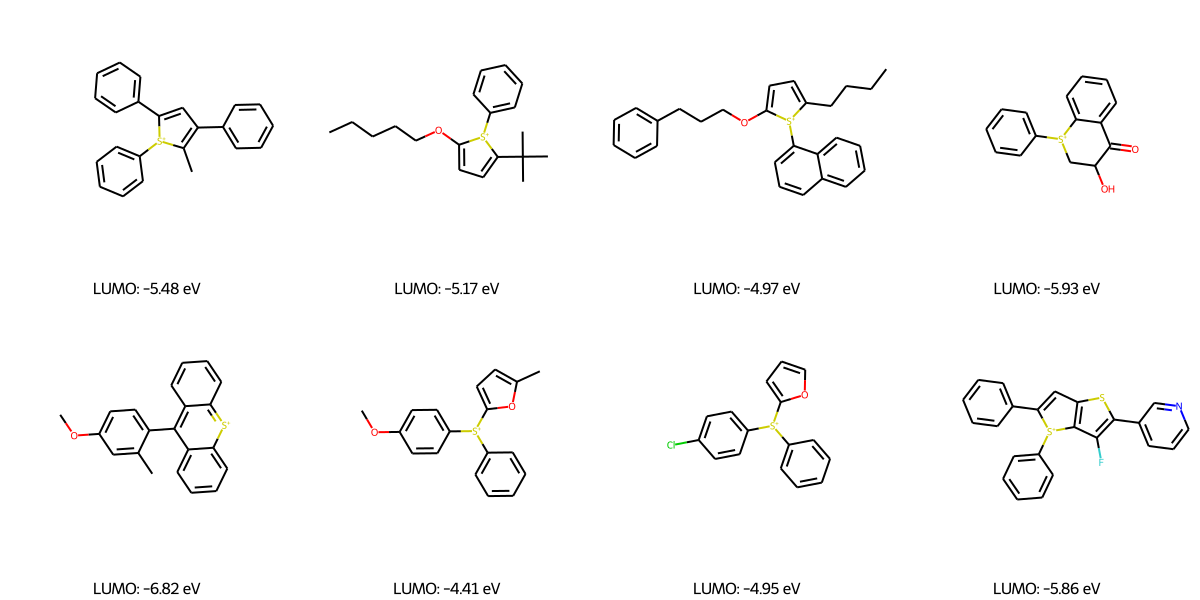}
    \caption{Molecules containing ``accepted'' scaffolds as evaluated by SME.}
    \label{fig:accepted_lumo}
\end{figure}

\begin{figure}[ht]
    \centering
    \includegraphics[height=\textheight]{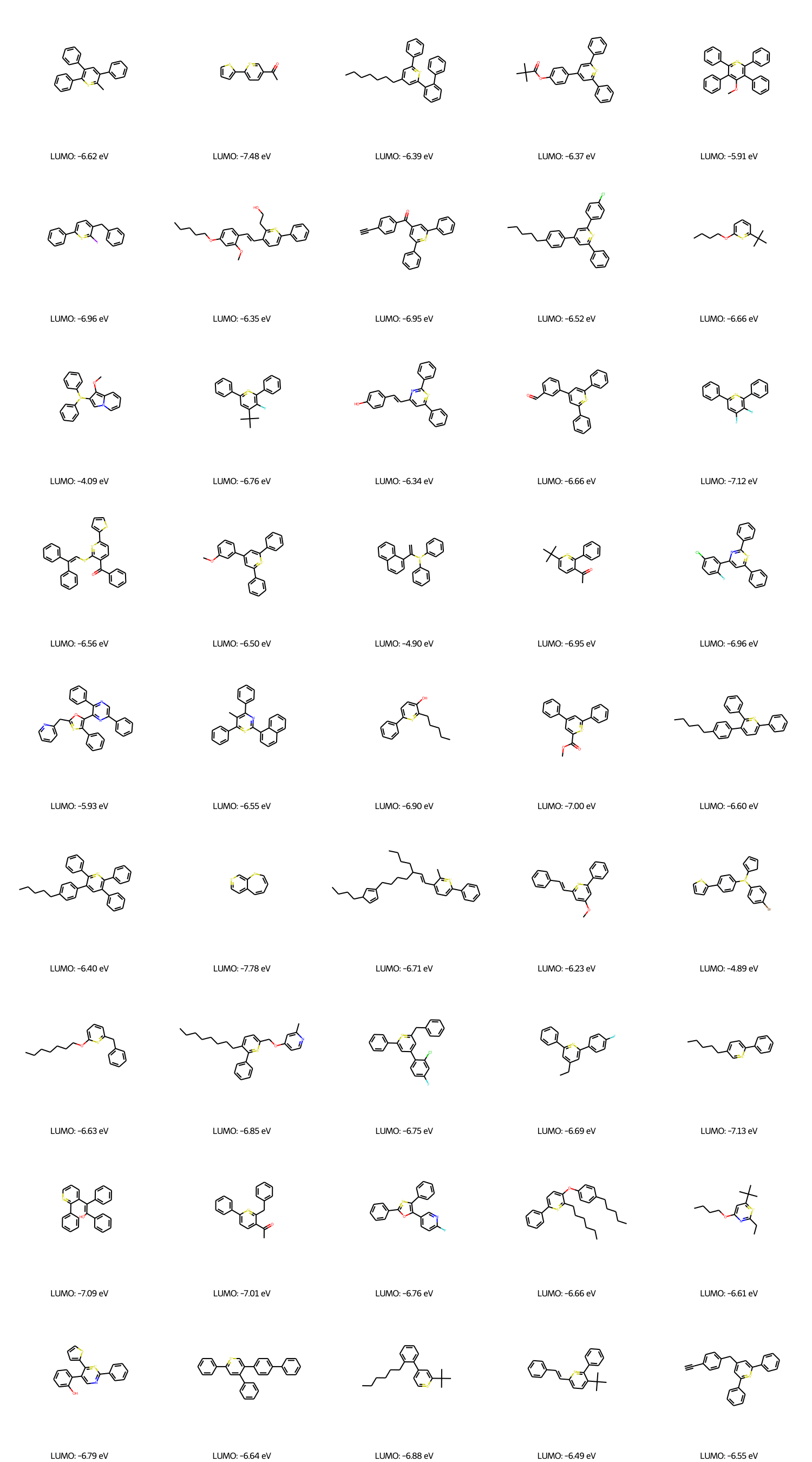}
    \caption{Molecules containing ``uncertain'' scaffolds as evaluated by SME.}
    \label{fig:uncertain_lumo}
\end{figure}

\end{document}

%% file: math_commands.tex
\usepackage{amsmath,amsfonts,bm}

\def\eqref#1{equation~\ref{#1}}

\def\1{\bm{1}}

\def\rva{{\mathbf{a}}}

\def\rvx{{\mathbf{x}}}

\def\rvz{{\mathbf{z}}}

\DeclareMathAlphabet{\mathsfit}{\encodingdefault}{\sfdefault}{m}{sl}
\SetMathAlphabet{\mathsfit}{bold}{\encodingdefault}{\sfdefault}{bx}{n}

\newcommand{\E}{\mathbb{E}}

\newcommand{\R}{\mathbb{R}}




%% file: main.bbl
\begin{thebibliography}{10}
\providecommand{\url}[1]{#1}
\csname url@samestyle\endcsname
\providecommand{\newblock}{\relax}
\providecommand{\bibinfo}[2]{#2}
\providecommand{\BIBentrySTDinterwordspacing}{\spaceskip=0pt\relax}
\providecommand{\BIBentryALTinterwordstretchfactor}{4}
\providecommand{\BIBentryALTinterwordspacing}{\spaceskip=\fontdimen2\font plus
\BIBentryALTinterwordstretchfactor\fontdimen3\font minus
  \fontdimen4\font\relax}
\providecommand{\BIBforeignlanguage}[2]{{%
\expandafter\ifx\csname l@#1\endcsname\relax
\typeout{** WARNING: IEEEtran.bst: No hyphenation pattern has been}%
\typeout{** loaded for the language `#1'. Using the pattern for}%
\typeout{** the default language instead.}%
\else
\language=\csname l@#1\endcsname
\fi
#2}}
\providecommand{\BIBdecl}{\relax}
\BIBdecl

\bibitem{PAG_History_1999}
J.~V. Crivello, ``The discovery and development of onium salt cationic
  photoinitiators,'' \emph{Journal of Polymer Science Part A: Polymer
  Chemistry}, vol.~37, no.~23, pp. 4241--4254, 1999.

\bibitem{PAG_Design_2019}
N.~Zivic, P.~K. Kuroishi, F.~Dumur, D.~Gigmes, A.~P. Dove, and H.~Sardon,
  ``Recent advances and challenges in the design of organic photoacid and
  photobase generators for polymerizations,'' \emph{Angewandte Chemie
  International Edition}, vol.~58, no.~31, pp. 10\,410--10\,422, 2019.

\bibitem{PAG_Env_2021}
X.-Z. Niu, R.~D. Pepel, R.~Paniego, J.~A. Field, J.~Chorover, L.~Abrell, A.~E.
  Sáez, and R.~Sierra-Alvarez, ``Photochemical fate of sulfonium photoacid
  generator cations under photolithography relevant uv irradiation,''
  \emph{Journal of Photochemistry and Photobiology A: Chemistry}, vol. 416, p.
  113324, 2021.

\bibitem{PAG_Env_2021A}
X.~Niu, J.~Field, R.~Paniego, R.~Pepel, J.~Chorover, L.~Abrell, and
  R.~Sierra-Alvarez, ``\BIBforeignlanguage{English (US)}{Bioconcentration
  potential and microbial toxicity of onium cations in photoacid generators},''
  \emph{\BIBforeignlanguage{English (US)}{Environmental Science and Pollution
  Research}}, vol.~28, no.~7, pp. 8915--8921, Feb. 2021, funding Information:
  This study was funded by the Semiconductor Industry Association (SIA) and
  Semiconductor Research Corporation (SRC). UA/NASA Undergraduate Internship
  Program is acknowledged for providing partial support for RDP. Publisher
  Copyright: {\textcopyright} 2021, The Author(s), under exclusive licence to
  Springer-Verlag GmbH, DE part of Springer Nature.

\bibitem{PAG_Design_2011}
Y.~Cho, C.~Y. Ouyang, M.~Krysak, W.~Sun, V.~Gamez, R.~Sierra-Alvarez, and C.~K.
  Ober, ``{Environmentally friendly natural materials-based photoacid
  generators for next-generation photolithography},'' in \emph{Advances in
  Resist Materials and Processing Technology XXVIII}, R.~D. Allen and M.~H.
  Somervell, Eds., vol. 7972, International Society for Optics and
  Photonics.\hskip 1em plus 0.5em minus 0.4em\relax SPIE, 2011, pp. 626 -- 636.

\bibitem{PAG_General_2020}
N.~A. Kuznetsova, G.~V. Malkov, and B.~G. Gribov, ``Photoacid generators.
  application and current state of development,'' \emph{Russian Chemical
  Reviews}, vol.~89, no.~2, pp. 173--190, jan 2020.

\bibitem{polishchuk2013estimation}
P.~G. Polishchuk, T.~I. Madzhidov, and A.~Varnek, ``Estimation of the size of
  drug-like chemical space based on {GDB}-17 data,'' \emph{Journal of
  Computer-Aided Molecular Design}, vol.~27, no.~8, pp. 675--679, 2013.

\bibitem{reutlinger2014multi}
M.~Reutlinger, T.~Rodrigues, P.~Schneider, and G.~Schneider, ``Multi-objective
  molecular de novo design by adaptive fragment prioritization,''
  \emph{Angewandte Chemie International Edition}, vol.~53, no.~16, pp.
  4244--4248, 2014.

\bibitem{yuan2011ligbuilder}
Y.~Yuan, J.~Pei, and L.~Lai, ``Ligbuilder 2: a practical de novo drug design
  approach,'' \emph{Journal of chemical information and modeling}, vol.~51,
  no.~5, pp. 1083--1091, 2011.

\bibitem{nigam2019augmenting}
A.~Nigam, P.~Friederich, M.~Krenn, and A.~Aspuru-Guzik, ``Augmenting genetic
  algorithms with deep neural networks for exploring the chemical space,'' in
  \emph{International Conference on Learning Representations}, 2020.

\bibitem{winter2019efficient}
R.~Winter, F.~Montanari, A.~Steffen, H.~Briem, F.~No{\'e}, and D.-A. Clevert,
  ``Efficient multi-objective molecular optimization in a continuous latent
  space,'' \emph{Chemical science}, vol.~10, no.~34, pp. 8016--8024, 2019.

\bibitem{EITL_2020}
\BIBentryALTinterwordspacing
P.~Ristoski, D.~Y. Zubarev, A.~L. Gentile, N.~Park, D.~Sanders, D.~Gruhl,
  L.~Kato, and S.~Welch, ``Expert-in-the-loop ai for polymer discovery,'' in
  \emph{Proceedings of the 29th ACM International Conference on Information and
  Knowledge Management}, ser. CIKM 20.\hskip 1em plus 0.5em minus 0.4em\relax
  New York, NY, USA: Association for Computing Machinery, 2020, p. 2701–2708.
  [Online]. Available: \url{https://doi.org/10.1145/3340531.3416020}
\BIBentrySTDinterwordspacing

\bibitem{segler2017generating}
M.~H. Segler, T.~Kogej, C.~Tyrchan, and M.~P. Waller, ``Generating focused
  molecule libraries for drug discovery with recurrent neural networks,''
  \emph{ACS Central Science}, vol.~4, no.~1, pp. 120--131, 2017.

\bibitem{gupta2018generative}
A.~Gupta, A.~T. M{\"u}ller, B.~J. Huisman, J.~A. Fuchs, P.~Schneider, and
  G.~Schneider, ``Generative recurrent networks for de novo drug design,''
  \emph{Molecular Informatics}, vol.~37, no. 1-2, p. 1700111, 2018.

\bibitem{guimaraes2017objective}
G.~L. Guimaraes, B.~Sanchez-Lengeling, C.~Outeiral, P.~L.~C. Farias, and
  A.~Aspuru-Guzik, ``Objective-reinforced generative adversarial networks
  ({ORGAN}) for sequence generation models,'' \emph{arXiv:1705.10843}, 2017.

\bibitem{de2018molgan}
N.~De~Cao and T.~Kipf, ``{MolGAN}: An implicit generative model for small
  molecular graphs,'' \emph{arXiv:1805.11973}, 2018.

\bibitem{gomez2018automatic}
R.~G{\'o}mez-Bombarelli, J.~N. Wei, D.~Duvenaud, J.~M. Hern{\'a}ndez-Lobato,
  B.~S{\'a}nchez-Lengeling, D.~Sheberla, J.~Aguilera-Iparraguirre, T.~D.
  Hirzel, R.~P. Adams, and A.~Aspuru-Guzik, ``Automatic chemical design using a
  data-driven continuous representation of molecules,'' \emph{ACS Central
  Science}, vol.~4, no.~2, pp. 268--276, 2018.

\bibitem{blaschke2018application}
T.~Blaschke, M.~Olivecrona, O.~Engkvist, J.~Bajorath, and H.~Chen,
  ``Application of generative autoencoder in de novo molecular design,''
  \emph{Molecular Informatics}, vol.~37, no. 1-2, p. 1700123, 2018.

\bibitem{kang2018conditional}
S.~Kang and K.~Cho, ``Conditional molecular design with deep generative
  models,'' \emph{Journal of Chemical Information and Modeling}, vol.~59,
  no.~1, pp. 43--52, 2018.

\bibitem{lim2018molecular}
J.~Lim, S.~Ryu, J.~W. Kim, and W.~Y. Kim, ``Molecular generative model based on
  conditional variational autoencoder for de novo molecular design,''
  \emph{Journal of Cheminformatics}, vol.~10, no.~1, p.~31, 2018.

\bibitem{popova2018deep}
M.~Popova, O.~Isayev, and A.~Tropsha, ``Deep reinforcement learning for de novo
  drug design,'' \emph{Science Advances}, vol.~4, no.~7, p. eaap7885, 2018.

\bibitem{olivecrona2017molecular}
M.~Olivecrona, T.~Blaschke, O.~Engkvist, and H.~Chen, ``Molecular de-novo
  design through deep reinforcement learning,'' \emph{Journal of
  Cheminformatics}, vol.~9, no.~1, p.~48, 2017.

\bibitem{jaques2017sequence}
N.~Jaques, S.~Gu, D.~Bahdanau, J.~M. Hern{\'a}ndez-Lobato, R.~E. Turner, and
  D.~Eck, ``Sequence tutor: Conservative fine-tuning of sequence generation
  models with {KL}-control,'' in \emph{Proceedings of the 34th ICML}.\hskip 1em
  plus 0.5em minus 0.4em\relax JMLR.org, 2017, pp. 1645--1654.

\bibitem{putin2018reinforced}
E.~Putin, A.~Asadulaev, Y.~Ivanenkov, V.~Aladinskiy, B.~Sanchez-Lengeling,
  A.~Aspuru-Guzik, and A.~Zhavoronkov, ``Reinforced adversarial neural computer
  for de novo molecular design,'' \emph{Journal of Chemical Information and
  Modeling}, vol.~58, no.~6, pp. 1194--1204, 2018.

\bibitem{Zhavoronkov2019natbio}
A.~Zhavoronkov, Y.~A. Ivanenkov, A.~Aliper, M.~S. Veselov, V.~A. Aladinskiy,
  A.~V. Aladinskaya, V.~A. .~Terentiev, D.~A. Polykovskiy, M.~D. Kuznetsov,
  A.~Asadulaev, Y.~Volkov, A.~Zholus, R.~R. Shayakhmetov, A.~Zhebrak, L.~I.
  Minaeva, B.~A. Zagribelnyy, L.~H. Lee, R.~Soll, D.~Madge, L.~Xing, G.~Tao,
  and A.~Aspuru-Guzik, ``Deep learning enables rapid identification of potent
  {DDR1} kinase inhibitors,'' \emph{Nature Biotechnology}, vol.~37, pp.
  1038--–1040, 2019.

\bibitem{born2020paccmann}
J.~Born, M.~Manica, A.~Oskooei, J.~Cadow, and M.~R. Mart{\'\i}nez,
  ``{PaccMann\textsuperscript{RL}}: Designing anticancer drugs from
  transcriptomic data via reinforcement learning,'' in \emph{International
  Conference on Research in Computational Molecular Biology}.\hskip 1em plus
  0.5em minus 0.4em\relax Springer, 2020, pp. 231--233.

\bibitem{zhou2019optimization}
Z.~Zhou, S.~Kearnes, L.~Li, R.~N. Zare, and P.~Riley, ``Optimization of
  molecules via deep reinforcement learning,'' \emph{Scientific Reports},
  vol.~9, no.~1, p. 10752, 2019.

\bibitem{li2018multi}
Y.~Li, L.~Zhang, and Z.~Liu, ``Multi-objective de novo drug design with
  conditional graph generative model,'' \emph{Journal of Cheminformatics},
  vol.~10, no.~1, p.~33, 2018.

\bibitem{das2020science}
P.~Das, T.~Sercu, K.~Wadhawan, I.~Padhi, S.~Gehrmann, F.~Cipcigan,
  V.~Chenthamarakshan, H.~Strobelt, C.~d. Santos, P.-Y. Chen \emph{et~al.},
  ``Accelerating antimicrobial discovery with controllable deep generative
  models and molecular dynamics,'' \emph{arXiv:2005.11248}, 2020.

\bibitem{chenthamarakshan2020cogmol}
\BIBentryALTinterwordspacing
V.~Chenthamarakshan, P.~Das, S.~Hoffman, H.~Strobelt, I.~Padhi, K.~W. Lim,
  B.~Hoover, M.~Manica, J.~Born, T.~Laino, and A.~Mojsilovic, ``Cogmol:
  Target-specific and selective drug design for covid-19 using deep generative
  models,'' in \emph{Advances in Neural Information Processing Systems},
  H.~Larochelle, M.~Ranzato, R.~Hadsell, M.~F. Balcan, and H.~Lin, Eds.,
  vol.~33.\hskip 1em plus 0.5em minus 0.4em\relax Curran Associates, Inc.,
  2020, pp. 4320--4332. [Online]. Available:
  \url{https://proceedings.neurips.cc/paper/2020/file/2d16ad1968844a4300e9a490588ff9f8-Paper.pdf}
\BIBentrySTDinterwordspacing

\bibitem{GAMESS}
\BIBentryALTinterwordspacing
G.~M.~J. Barca, C.~Bertoni, L.~Carrington, D.~Datta, N.~De~Silva, J.~E.
  Deustua, D.~G. Fedorov, J.~R. Gour, A.~O. Gunina, E.~Guidez, T.~Harville,
  S.~Irle, J.~Ivanic, K.~Kowalski, S.~S. Leang, H.~Li, W.~Li, J.~J. Lutz,
  I.~Magoulas, J.~Mato, V.~Mironov, H.~Nakata, B.~Q. Pham, P.~Piecuch,
  D.~Poole, S.~R. Pruitt, A.~P. Rendell, L.~B. Roskop, K.~Ruedenberg,
  T.~Sattasathuchana, M.~W. Schmidt, J.~Shen, L.~Slipchenko, M.~Sosonkina,
  V.~Sundriyal, A.~Tiwari, J.~L. Galvez~Vallejo, B.~Westheimer, M.~Wloch,
  P.~Xu, F.~Zahariev, and M.~S. Gordon, ``\BIBforeignlanguage{en}{Recent
  developments in the general atomic and molecular electronic structure
  system},'' \emph{\BIBforeignlanguage{en}{The Journal of Chemical Physics}},
  vol. 152, no.~15, p. 154102, Apr. 2020. [Online]. Available:
  \url{http://aip.scitation.org/doi/10.1063/5.0005188}
\BIBentrySTDinterwordspacing

\bibitem{zinc}
\BIBentryALTinterwordspacing
T.~Sterling and J.~J. Irwin, ``Zinc 15 – ligand discovery for everyone,''
  \emph{Journal of Chemical Information and Modeling}, vol.~55, no.~11, pp.
  2324--2337, 2015, pMID: 26479676. [Online]. Available:
  \url{https://doi.org/10.1021/acs.jcim.5b00559}
\BIBentrySTDinterwordspacing

\bibitem{moses}
D.~Polykovskiy, A.~Zhebrak, B.~Sanchez-Lengeling, S.~Golovanov, O.~Tatanov,
  S.~Belyaev, R.~Kurbanov, A.~Artamonov, V.~Aladinskiy, M.~Veselov, A.~Kadurin,
  S.~Johansson, H.~Chen, S.~Nikolenko, A.~Aspuru-Guzik, and A.~Zhavoronkov,
  ``{M}olecular {S}ets ({MOSES}): {A} {B}enchmarking {P}latform for {M}olecular
  {G}eneration {M}odels,'' \emph{Frontiers in Pharmacology}, 2020.

\bibitem{degen2008}
\BIBentryALTinterwordspacing
J.~Degen, C.~Wegscheid-Gerlach, A.~Zaliani, and M.~Rarey, ``On the art of
  compiling and using {\textquotesingle}drug-like{\textquotesingle} chemical
  fragment spaces,'' \emph{{ChemMedChem}}, vol.~3, no.~10, pp. 1503--1507, Oct.
  2008. [Online]. Available: \url{https://doi.org/10.1002/cmdc.200800178}
\BIBentrySTDinterwordspacing

\bibitem{murckoscaf}
\BIBentryALTinterwordspacing
G.~W. Bemis and M.~A. Murcko, ``The properties of known drugs. 1. molecular
  frameworks,'' \emph{Journal of Medicinal Chemistry}, vol.~39, no.~15, pp.
  2887--2893, 1996, pMID: 8709122. [Online]. Available:
  \url{https://doi.org/10.1021/jm9602928}
\BIBentrySTDinterwordspacing

\end{thebibliography}
